\documentclass[letterpaper]{article} %
\usepackage{aaai25}  %
\usepackage{times}  %
\usepackage{helvet}  %
\usepackage{courier}  %
\usepackage[hyphens]{url}  %
\usepackage{graphicx} %
\usepackage{natbib}  %
\usepackage{caption} %
\usepackage{algorithm}
\usepackage{algorithmic}
\usepackage[dvipsnames]{xcolor}
\usepackage[most]{tcolorbox}
\usepackage[T1]{fontenc}

\definecolor{prompt}{HTML}{5f84e4}
\definecolor{img}{HTML}{820100}

\usepackage{newfloat}
\usepackage{listings}
\DeclareCaptionStyle{ruled}{labelfont=normalfont,labelsep=colon,strut=off} %
\floatstyle{ruled}
\newfloat{listing}{tb}{lst}{}
\floatname{listing}{Listing}
\usepackage{abbrv}
\usepackage{amsfonts}
\usepackage{amsmath}
\usepackage{booktabs}
\usepackage{multirow}
\usepackage[capitalize]{cleveref}
\Crefname{figure}{Figure}{Figures}
\newcommand{\correct}[1]{\text{\color{Green}{\underline{#1}}}}
\newcommand{\wrong}[1]{\text{\color{Red}{\underline{#1}}}}

\newcommand{\R}{\mathbb{R}}
\def\method{\textsc{CAD2Program}\xspace}

\title{From 2D CAD Drawings to 3D Parametric Models: A Vision-Language Approach}

\author{
Xilin Wang\textsuperscript{\rm 1}\equalcontrib,
Jia Zheng\textsuperscript{\rm 2}\equalcontrib,
Yuanchao Hu\textsuperscript{\rm 2},
Hao Zhu\textsuperscript{\rm 2},
Qian Yu\textsuperscript{\rm 1}$^{\dagger}$,
Zihan Zhou\textsuperscript{\rm 2}$^{\dagger}$
}

\affiliations{
\textsuperscript{\rm 1}Beihang University, 
\textsuperscript{\rm 2}Manycore Tech Inc.\\
\{wang\_xilin, qianyu\}@buaa.edu.cn, \{jiajia, tiancai, hunyuan, shuer\}@qunhemail.com \\
\url{https://manycore-research.github.io/CAD2Program}
}

\begin{document}

\maketitle

\begin{abstract}

In this paper, we present \method, a new method for reconstructing 3D parametric models from 2D CAD drawings. Our proposed method is inspired by recent successes in vision-language models (VLMs), and departs from traditional methods which rely on task-specific data representations and/or algorithms. Specifically, on the input side, we simply treat the 2D CAD drawing as a raster image, regardless of its original format, and encode the image with a standard ViT model. We show that such an encoding scheme achieves competitive performance against existing methods that operate on vector-graphics inputs, while imposing substantially fewer restrictions on the 2D drawings. On the output side, our method auto-regressively predicts a general-purpose language describing 3D parametric models in text form. Compared to other sequence modeling methods for CAD which use domain-specific sequence representations with fixed-size slots, our text-based representation is more flexible, and can be easily extended to arbitrary geometric entities and semantic or functional properties. Experimental results on a large-scale dataset of cabinet models demonstrate the effectiveness of our method.

\end{abstract}

\section{Introduction}

In computer-aided design (CAD), 2D engineering drawings have been employed as a standard means for describing product designs. With the wide adoption of 2D CAD software (\eg, Autodesk AutoCAD), a significant amount of existing products are presented in the form of engineering drawings today. To manufacture these products, the corresponding 3D objects have to be reconstructed from the 2D drawings.

\begin{figure*}[t]
    \centering
    \includegraphics[width=0.99\linewidth]{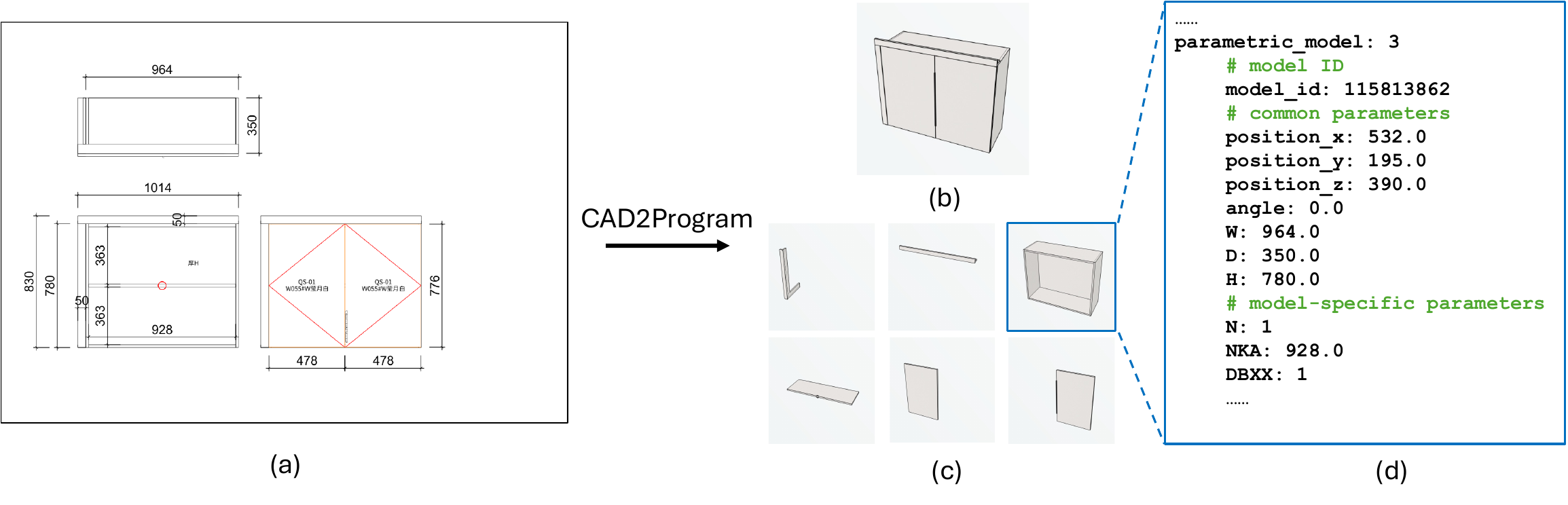}    
    \caption{Problem statement. Given {\bf (a)} a 2D CAD drawing of a product (\eg, a cabinet), our goal is to reconstruct {\bf (b)} the 3D model of a product. {\bf (c)} In CAD software, the 3D model is conventionally built by assembling pre-defined primitive models, where {\bf (d)} each primitive model is defined by a computer program describing its model ID and a number of parameters.}
    \label{fig:pipeline}
\end{figure*}

As shown in \cref{fig:pipeline}(a), an engineering drawing typically includes multiple orthographic views of the object. Each orthographic view is a projection of the object onto the plane that is perpendicular to one of the three principal axes.\footnote{Depending on the complexity of the object, a varying number of views may be provided, including a top view, a front view, a side view, and one or more section views.}
A line of work related to the automatic reconstruction of 3D solid models from orthographic views has existed since the 1970s~\cite{IdesawaM73}. However, in the current design and manufacturing industry, human labor is still extensively used to manually realize these 3D object models, which is a time-consuming, error-prone, and unproductive process. We are yet to see the successful application of automatic techniques in commercial CAD software. 

To understand the challenges faced by existing automatic techniques, we must take a close look at (i) the engineering drawings and (ii) the 3D models.

\begin{figure}[t]
    \centering
    \includegraphics[width=0.9\linewidth]{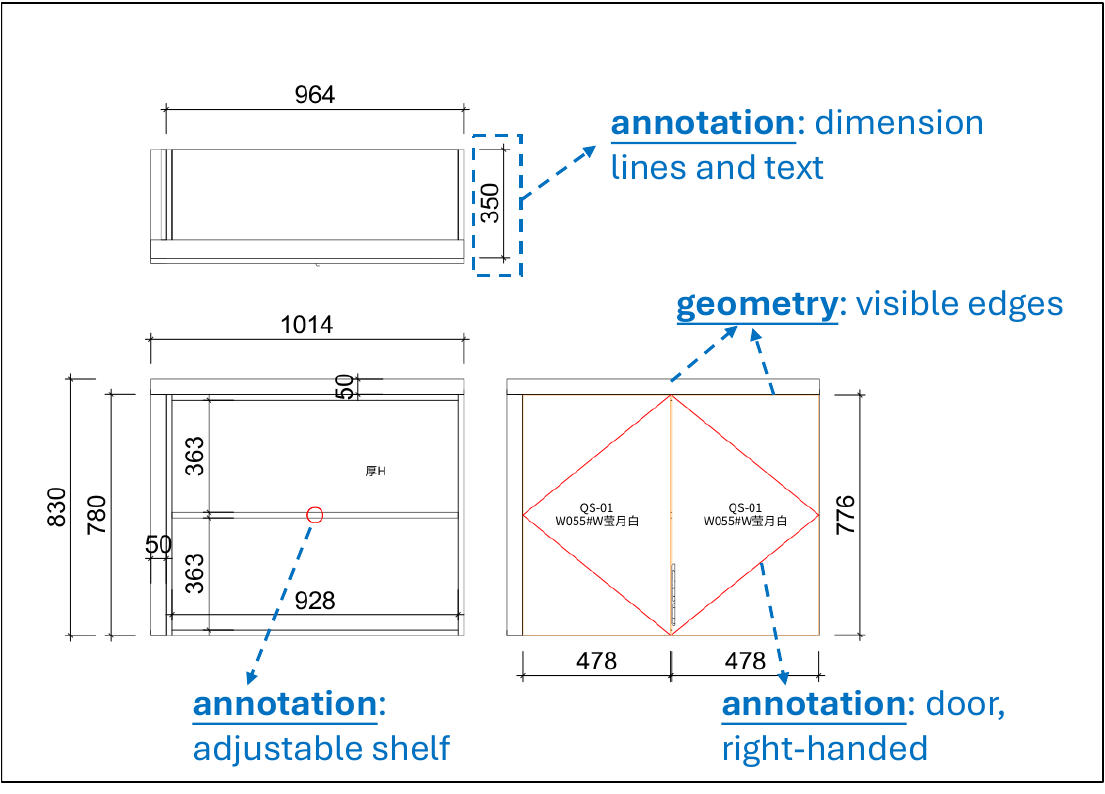}
    \caption{Illustration of the geometry and annotation layers of a CAD drawing. See text for details.}     
    \label{fig:layers} 
\end{figure}

\smallskip
\noindent\textbf{Understanding the 2D drawings.} An engineering drawing poses a unique challenge as it is a mixture of two types of representations (layers):
\begin{itemize}
    \item the \emph{geometry layer}, which is the actual object described by its orthographic projections, and
    \item the \emph{annotation layer}, which includes dimensioning and functional symbols, such as surface types, manufacturing instructions, \etc
\end{itemize}
\cref{fig:layers} illustrates the two layers. Some lines constitute the geometry layer of the drawing, as they correspond to the visible edges of the object in the orthographic projection. The other entities, including both text and graphics, form the annotation layer. For example, many of the numbers, together with the associated lines, specify the precise sizes (width, depth, and height) of the object parts. A red circle denotes that the shelf is adjustable, whereas the red triangle indicates a door and its opening direction.

Most existing methods for 3D reconstruction from engineering drawings focus on the \emph{geometry layer only}, \ie, to reconstruct 3D B-rep or CSG models by matching the boundary projections across orthographic views~\cite{SakuraiG83, GuTS86, LequetteR88, YanCT94, YouY96, ShinS98, Kuo98, LiuHCS01, GongZZS06a, GongZZS06b}. They all assume that a clean geometry layer with exactly three axis-aligned views (namely, front, top, and side views) is provided, and any ``irrelevant'' annotation has been removed. In practice, this assumption is problematic for at least two reasons: \emph{First}, separating the annotation from the geometry layer is a non-trivial task. As one can see in \cref{fig:layers}, some lines correspond to the visible edges of the object and belong to the geometry layer, while others belong to the annotation layer (\eg, dimension sets). Thus, the customary text/graphics layer separation does not work here. \emph{Second}, by omitting the annotations, important information about the 3D object, such as dimensions and functional symbols, is ignored during the reconstruction process.

\smallskip
\noindent\textbf{Understanding the 3D models.} Now we turn our attention to the output. In CAD software, B-rep is a universal data format, which represents a 3D object as a volume contained in a set of parametric surfaces together with the relationships between them (\ie, the topology). However, human designers rarely create product designs by directly working with these low-level geometric entities. Instead, they make use of pre-defined primitive models (or ``parts'') to (i) speed up the design process, and (ii) reduce design errors and ensure that the product can be properly manufactured.

\begin{figure}[t]
    \centering
    \includegraphics[width=0.99\linewidth]{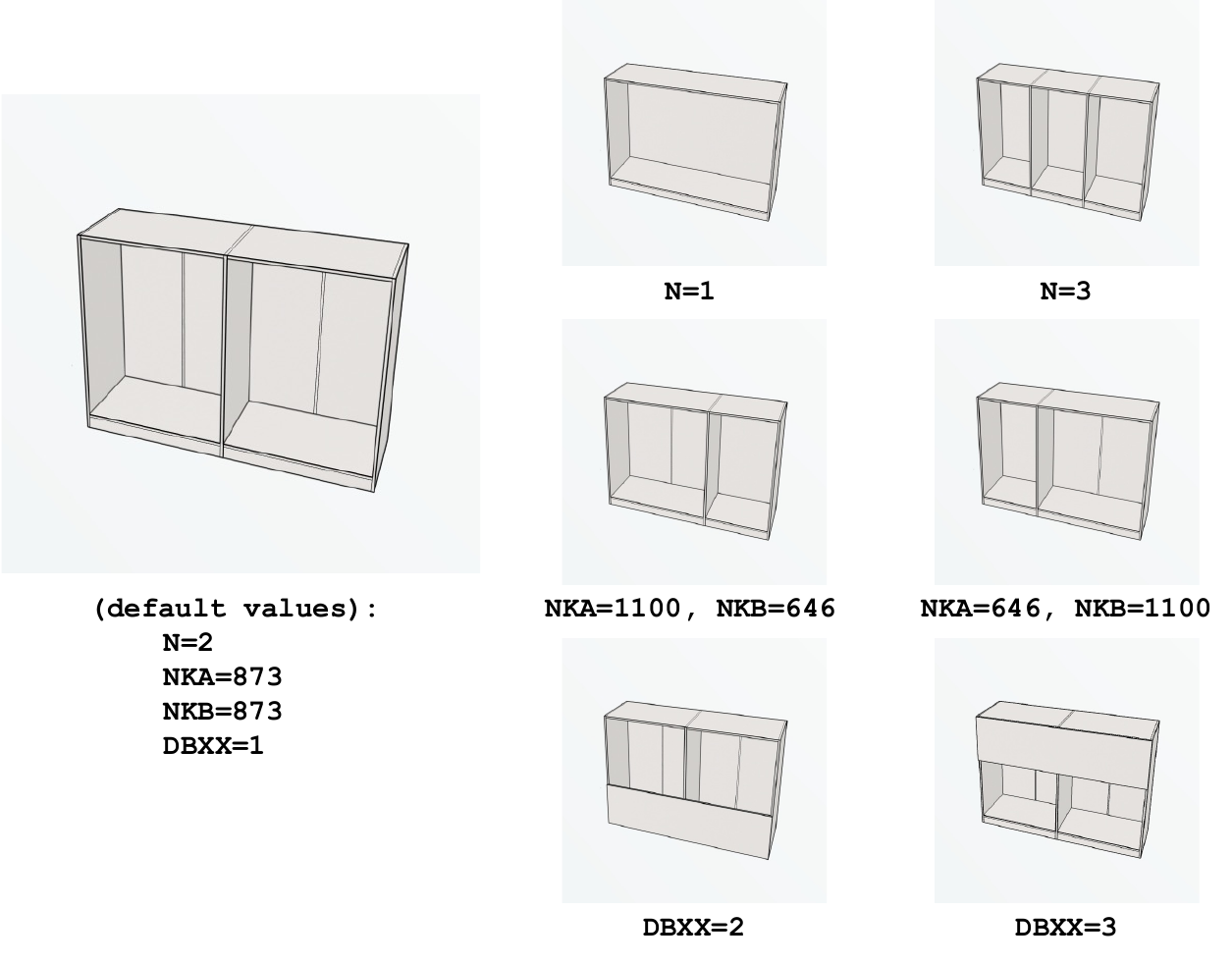}    
    \caption{Illustration of the model-specific parameters of a ``base box'' primitive. \texttt{N} is the number of vertically divided spaces in the box. [\texttt{NKA}, \texttt{NKB}, $\ldots$] are the widths of the divided spaces. \texttt{DBXX} indicates the position of the frame, where \texttt{DBXX=1} means ``no frame'', \texttt{DBXX=2} means ``lower frame'', and \texttt{DBXX=3} means ``upper frame''.} 
    \label{fig:parameters} 
\end{figure}

Consider a specific type of product, namely the \emph{cabinet furniture}, for example. Some common primitives may include ``base box'', ``door'', ``drawer'', ``fixed shelf'', `adjustable shelf'', and so on. \cref{fig:pipeline}(c) shows an example of decomposing a simple cabinet into the primitives. As one can see in Figure~\ref{fig:pipeline}(d), each primitive itself is a parametric model. To use a primitive in the product design, one must specify the following:
\begin{itemize}
    \item the \emph{model ID}, which is the unique identifier of a primitive in the database,
    \item the \emph{common parameters}, which indicate the general pose and size of the primitive in the 3D space and are model-agnostic, and
    \item the \emph{model-specific parameters}, which describe possible variations of a specific primitive (see Figure~\ref{fig:parameters} for a visualization).
\end{itemize}

Since a parametric model can be specified by a list of parameters, a line of recent studies~\cite{NashGEB20, JonesBX0JGMR20, WuXZ21, WillisJLCP21, GaninBLKS21, SeffZRA22, GuoLPLTG22, XuWLCJF22, YuCTMZ23, JayaramanLDWSM23, XuJLWF23} leverage Transformer-based sequence modeling techniques to auto-regressively predict these parameters for CAD model generation. To this end, they define domain-specific sequence representations by collecting all possible parameters and assigning a fixed slot to each parameter in the sequence. One issue with these representations is that, due to the presence of model-specific parameters, the number of slots needed grows with the set of primitives. As a result, these methods typically deal with a small set of (\eg, $\leq 10$) primitives.

\smallskip
\noindent\textbf{Our contributions.} In this paper, we address the aforementioned challenges as follows.

\emph{First}, on the input side, instead of extracting and analyzing the graphic elements in the geometry layer as existing methods do, we advocate a more holistic approach in which we simply treat the entire 2D drawing as a raster image and process it with an image encoder (\eg, a ViT). This way, our method not only makes use of information in both geometry and annotation layers for reconstruction, but also eliminates the requirements imposed by existing methods such as a separate geometry layer being available, or exactly three orthographic views (\ie, front, top, and side) are present. 

\emph{Second}, on the output side, we represent 3D parametric models as scripts of a general-purpose language (\eg, Python) and auto-regressively predict the language in text form. Our design choice offers several advantages: (i) it is interpretable and semantically rich; (ii) it can seamlessly integrate new primitives without affecting the length of existing scripts; and (iii) it allows us to leverage strong built-in coding capability of pre-trained large language models.

\emph{Finally}, with the above input-output setting, we implement our method, \method, by directly fine-tuning an open-source vision-language foundation model. In this paper, we choose InternVL~\cite{InternVL1.5, InternVL} as the base model. To validate our design choices, we have collected a dataset consisting of 368K cabinet models with 2D engineering drawings. Comprehensive experiments on the dataset demonstrate the effectiveness of our method.

\section{Related Work}

\subsection{3D Reconstruction from Orthographic Views} 

The problem of 3D reconstructing from orthographic views is a long-standing problem in CAD~\cite{WangG93}. \cite{IdesawaM73} and \cite{MarkowskyW80, WesleyM81} are the first to present a principled mathematical framework to this problem, and propose to recover 3D vertices, edges, faces, and solids progressively from the input views. Subsequent studies~\cite{SakuraiG83, GuTS86, LequetteR88, YanCT94, YouY96, ShinS98, Kuo98, LiuHCS01, GongZZS06a, GongZZS06b} largely follow the same scheme and focus on (i) improving the computational efficiency of existing algorithms and (ii) extending the applicability of the framework to handle more complex objects (\eg, quadric surface solids). 

Recently, a couple of learning-based approaches emerge as an alternative to the above scheme~\cite{HanXLWF20, HuZZYYZ23}. The most relevant work to ours is PlankAssembly~\cite{HuZZYYZ23}, which also aims to reconstruct 3D parametric models from 2D orthographic views. It employs an encoder-decoder architecture to learn an implicit mapping of geometric entities between the 3D model and 2D views. The main advantage of PlankAssembly is its robustness against errors and missing components in 2D drawings.

However, PlankAssembly has several limitations. \emph{First}, on the input side, it requires exactly three axis-aligned views in vector-graphics format as input, and encodes each entity (\ie, 2D line) in the geometry layer as a token. \emph{Second}, on the output side, it deals with a single type of primitive, namely cuboid-shaped planks with six parameters.

\subsection{Sequence Modeling for CAD} 

With the success of Transformer-based sequence modeling techniques~\cite{VaswaniSPUJGKP17} in the NLP field, a line of work~\cite{NashGEB20, JonesBX0JGMR20, WuXZ21, WillisJLCP21, GaninBLKS21, SeffZRA22, GuoLPLTG22, XuWLCJF22, YuCTMZ23, JayaramanLDWSM23, XuJLWF23} propose to design domain-specific languages (DSL) for the shapes of interest and train deep networks to generate CAD models auto-regressively. Other work~\cite{GuoLPLTG22, MaCLLZ24, LiGLBY24, CAD-SIGNet} reconstructs CAD construction sequences from voxels or point clouds.

As discussed before, while most existing methods learn domain-specific sequence representations, we represent 3D parametric models as text scripts using a general-purpose language. As a result, our method can efficiently deal with a large number of primitives and model-specific parameters.

\section{Method}

In this paper, we focus on cabinet furniture, a common type of parametric models, as our subject of study. We first introduce our text-based shape program in~\cref{sec:program}, then show how to train vision-language models to reconstruct 3D cabinet models from 2D drawings in~\cref{sec:model}.

\subsection{Cabinet Shape Program}
\label{sec:program}

As shown in Figure~\ref{fig:pipeline}, a 3D cabinet model is built by assembling a list of primitive instances $\mathcal{Z} = \{ Z_i \}_{i=1}^{N}$. Each primitive instance can be represented as a tuple $Z_i = \{ M_i, B_i, P_i\}$, where $M_i$ is a unique ID of the primitive model in the database, $B_i = \{ p_i, s_i, r_i \}$ defines a 3D box comprising all the common parameters including the center position $p_i \in \R^3$, size $s_i \in \R^3$, and 1-D rotation angle $r_i\in \R$ around the vertical axis\footnote{We do not use full 3D rotation because designers rarely rotate a model along the $x$- and $y$-axis.}, and $P_i = \{ \text{key}_j = \text{value}_j \}_{j=1}^K $ is a list of model-specific parameters associated with the primitive. Note that $P_i = \emptyset$ if the primitive $M_i$ has no model-specific parameters.

In this paper, we follow the convention of CAD and use a coordinate system in which the bounding box of the subject (\ie, cabinet) is aligned to the main axes, with its bottom face aligned with the $z$-axis, and its front face aligned with the $-y$-axis. The origin coincides with one of the corners of the box, such that the subject lies in the first octant.

\smallskip
\noindent\textbf{Text-based shape program.} A common practice for applying sequence modeling in the CAD field is to design domain-specific language (DSL) for the shape of interest, such as 2D sketches~\cite{WillisJLCP21, GaninBLKS21, SeffZRA22} and 3D solid models~\cite{WuXZ21, GuoLPLTG22, XuWLCJF22, JayaramanLDWSM23, XuJLWF23}. In these methods, all the parameters in the DSL are aggregated to form a fixed-length command template $C$. Then, the shape is represented as a sequence of commands $[C_1, C_2, C_3, \ldots]$. For example, DeepCAD~\cite{WuXZ21} deals with 3 types of geometric entities (\ie, lines, arcs, and circles) and 1 type of operation (\ie, extrusion) for 3D solid generation, resulting in a command template $C$ with 16 parameters. 

While a command template can represent arbitrarily complex models in theory, it becomes problematic when applied to our task: due to the presence of model-specific parameters, the length of the command template $C$ grows with the set of primitives. With hundreds or more primitives, the command sequence representing the 3D model would be prohibitively long, with a high percentage of unused slots in each command. 

\begin{listing}[t]
    \caption{Python shape program describing the cabinet in \cref{fig:pipeline}. Every two lines correspond to a primitive model in \cref{fig:pipeline}(c).}
    \label{lst:program}
\begin{lstlisting}[xleftmargin=0pt,numbers=none]
bbox_0 = Bbox(507, 185, 805, 1014, 370, 50, 0)
model_0 = <model_57761062>()
bbox_1 = Bbox(25, 185, 390, 50, 370, 780, 0)
model_1 = <model_57758898>()
bbox_2 = Bbox(532, 195, 390, 964, 350, 780, 0)
model_2 = <model_115813862>(N=1, NKA=928, DBXX=1, BT=18)
bbox_3 = Bbox(532, 185, 390, 928, 330, 18, 0)
model_3 = <model_57253481>()
bbox_4 = Bbox(291, 11, 390, 478, 18, 776, 0)
model_4 = <model_82289390>(openDirection=0, uCove=18, dCover=18, lCover=18, rCover=18)
bbox_5 = Bbox(773, 11, 390, 478, 18, 776, 0)
model_5 = <model_82289390>(openDirection=1, uCover=18, dCover=18, lCover=18, rCover=18)
\end{lstlisting}
\end{listing}

To this end, we find that these limitations can be addressed by directly representing each primitive in text form. In this paper, we choose Python as a proxy language and show an example of our text-based representation in \cref{lst:program}. Note that we include both the key and value of each model-specific parameter in the language. This way, we eliminate the need to construct a common command template. 

Besides its ability to handle an arbitrary number of primitives, we note that our Python-based shape program offers several other benefits: \emph{First}, in existing methods, the continuous value of a parameter in the command is quantized into discrete tokens by a domain-specific tokenizer, resulting in an inherent quantization error. By treating these values as text, the quantization error can be avoided. \emph{Second}, it allows us to leverage LLM's strong capacity to perform Python programming. \emph{Third}, the fact that our program is a sequence of text tokens opens up many potential applications such as text-based editing and visual question answering.

\subsection{The \method Model}
\label{sec:model}

Given the input-output settings, our task can be regarded as an application of vision-language foundation models (VLMs). These models aim to integrate and enhance LLM's capabilities in processing both visual and textual data. In the past year, significant progress has been made in developing both proprietary commercial models, such as OpenAI's GPT-4V~\cite{gpt-4v} and Google's Gemini-1.5~\cite{gemini1.5}, and open-source models including LLaVA~\cite{ImprovedLLaVA, LLaVA, LLaVA-NeXT, LLavA-Next-Strong}, MiniGPT-4~\cite{MiniGPT4}, InternVL~\cite{InternVL1.5, InternVL}, CogVLM~\cite{CogVLM}, and many more.

In this paper, we choose Mini-InternVL-1.5-2B for its open-source availability\footnote{\url{https://huggingface.co/OpenGVLab/Mini-InternVL-Chat-2B-V1-5}} and excellent balance between performance and model size. Notably, Mini-InternVL-1.5 supports dynamic resolution by splitting the high-resolution image into up to 12 tiles, significantly enhancing the performance on OCR-related tasks. Also, it is trained on high-quality bilingual dataset and has demonstrated robust capacities in handling multi-modal perception tasks in both English and Chinese. These characteristics make it well-suited for our task.

\begin{figure}[t]
    \centering
    \includegraphics[width=0.99\linewidth]{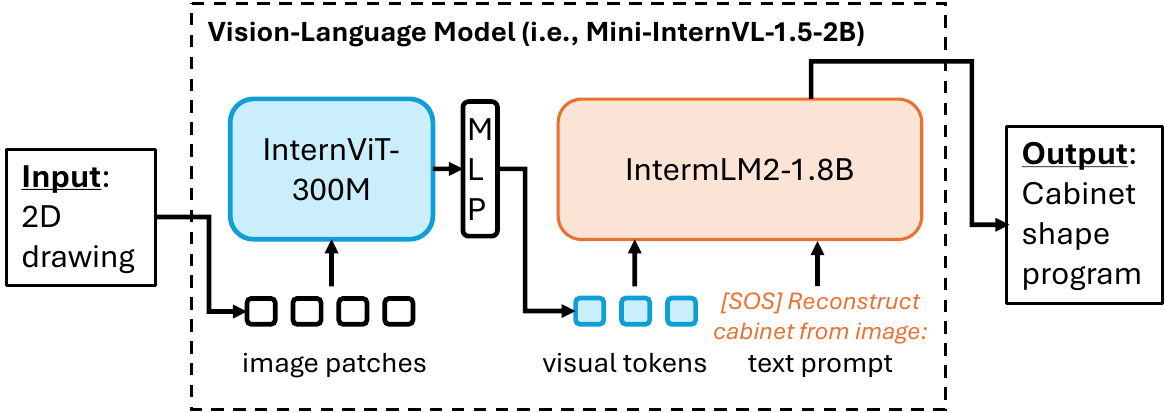}
    \caption{An overview of the \method model.} 
    \label{fig:internvl} 
\end{figure}

Figure~\ref{fig:internvl} provides an overview of our \method model. It adopts a ViT-MLP-LLM architecture, using InternViT-300M as the vision encoder and InternLM2-1.8B as the language model, respectively. These two models are aligned by a MLP projector. 
For our task, we use a simple phrase \emph{``Reconstruct cabinet from image:''} as the text prompt to the language model.

For the output, we note that the model ID is just a randomly generated number and does not carry any semantic meaning. To help the network better align the input visual tokens with the primitives in the database, we create a special token for the model ID. Specifically, we consider two features of the primitive: (i) the model name, and (ii) a snapshot image of the primitive rendered with its default model-specific parameter values from a fixed viewpoint. We use Chinese-CLIP~\cite{ChineseCLIP}\footnote{\url{https://huggingface.co/OFA-Sys/chinese-clip-vit-huge-patch14}} with a ViT-H/14 image encoder~\cite{ViT} and a RoBERTa-wwm-large text encoder~\cite{ChineseBert} to extract image and text embeddings, respectively. Then, the special token is obtained by directly concatenating these two embeddings.

\smallskip
\noindent \textbf{Dataset.} To train and test our model, we collect a new dataset by accessing a large repository of 3D customized cabinet models on an online interior design platform. On the platform, professional designers utilize 3D modeling software to create 3D cabinet models and the corresponding engineering drawings for real-world production.

\begin{figure}[t]
    \centering
    \includegraphics[width=0.99\linewidth]{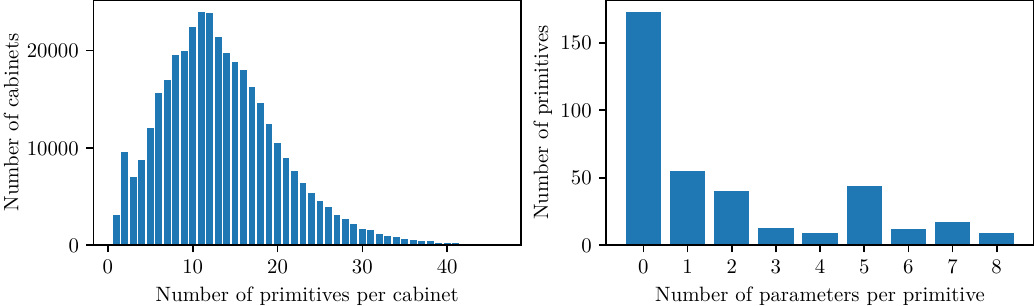}
    \caption{Dataset statistics. \textbf{Left:} the number of cabinets \wrt the number of primitives per cabinet. \textbf{Right:} the number of primitives \wrt the number of model-specific parameters per primitive.}
    \label{fig:statistics}
\end{figure}

When creating the dataset, we restrict the size of the cabinet to be between 0.1m and 4.5m to avoid undersized or oversized models, and also eliminate models with more than 48 primitives. After filtering, our dataset contains 368K cabinet models and 2D engineering drawings, with 373 unique pre-defined primitives. The number of model-specific parameters per primitive ranges from 0 to 8. The total number of model-specific parameters is 702 -- at least an order of magnitude larger than the number seen in any command template used in prior work. Some statistics of the dataset are shown in~\cref{fig:statistics}. Finally, the dataset is divided into 364K/2K/2K samples for training/validation/testing.

\smallskip
\noindent\textbf{Implementation details.} We use the SWIFT~\cite{Swift} framework to train \method via supervised full-parameter fine-tuning. We utilize the AdamW optimizer~\cite{LoshchilovH17} and a cosine learning rate schedule with a linear warm-up for 1K steps. The peak learning rate is $10^{-5}$. The model is trained for about 14K iterations, which takes about 1 day using 64 NVIDIA RTX 4090 GPU devices. The total batch size is set to 128. The length of the token sequence is restricted to 4096.

\section{Experiments}

\begin{listing*}[t]
    \captionof{listing}{Comparison of programs generated by models trained with or without the annotation layer. The corresponding input drawing can be found in \cref{fig:visuals} (first row, left). Due to the space limit, only the 3D box predictions are shown here. We use \underline{underline} to indicate numbers that can be directly referred from the dimension sets in the annotation layer. The green and red colors denote the correct and incorrect predictions, respectively. We reorder the predictions by matching them to the ground truth primitives. \textbf{Best viewed in color.}}
    \label{lst:layer}
    \begin{minipage}[t]{0.45\textwidth}
\begin{lstlisting}
# model trained with annotation layer
bbox_10 = Bbox(1637, 280, 1040, $\correct{50}$, 560, $\correct{2080}$, 0)
bbox_9 = Bbox(1637, 280, 2383, $\correct{50}$, 560, $\correct{606}$, 0)
bbox_0 = Bbox(806, 290, 1040, $\correct{1612}$, $\correct{540}$, $\correct{2080}$, 0)
bbox_3 = Bbox(407.5, 280, 1079, $\correct{779}$, 520, 18, 0)
bbox_4 = Bbox(407.5, 280, 978.5, $\correct{779}$, 40, 83, 0)
bbox_6 = Bbox(407.5, 280, 1740, $\correct{779}$, 520, 18, 0)
bbox_5 = Bbox(407.5, 280, 1409, $\correct{779}$, 520, 18, 0)
bbox_1 = Bbox(1204.5, 280, 1575, $\correct{779}$, 520, 18, 0)
bbox_2 = Bbox(1204.5, 280, 1474.5, $\correct{779}$, 40, 83, 0)
bbox_7 = Bbox(806, 290, 2383, $\correct{1612}$, $\correct{540}$, $\correct{606}$, 0)
bbox_8 = Bbox(831, 280, 2746, $\correct{1662}$, 560, $\correct{120}$, 0)
\end{lstlisting}
    \end{minipage}
    \hfill
    \begin{minipage}[t]{0.45\textwidth}
\begin{lstlisting}
# model trained without annotation layer
bbox_0 = Bbox(1665, 280, 1040, $\correct{50}$, 560, $\correct{2080}$, 0)
bbox_1 = Bbox(1665, 280, 2385, $\correct{50}$, 560, $\wrong{610}$, 0)
bbox_3 = Bbox(795, 290, 1040, $\wrong{1590}$, $\correct{540}$, $\correct{2080}$, 0)
bbox_4 = Bbox(402, 280, 1079, $\wrong{768}$, 520, 18, 0)
bbox_5 = Bbox(402, 280, 978.5, $\wrong{768}$, 40, 83, 0)
bbox_6 = Bbox(402, 280, 1740, $\wrong{768}$, 520, 18, 0)
bbox_7 = Bbox(402, 280, 1409, $\wrong{768}$, 520, 18, 0)
bbox_8 = Bbox(1188, 280, 1585, $\wrong{768}$, 520, 18, 0)
bbox_9 = Bbox(1188, 280, 1484.5, $\wrong{768}$, 40, 83, 0)
bbox_10 = Bbox(795, 290, 2385, $\wrong{1590}$, $\correct{540}$, $\wrong{610}$, 0)
bbox_2 = Bbox(820, 280, 2750, $\wrong{1640}$, 560, $\correct{120}$, 0)
\end{lstlisting}
    \end{minipage}
\end{listing*}

\subsection{Evaluation Metrics}

To evaluate our method, we consider the accuracy of (i) model retrieval (for model ID), (ii) 3D reconstruction (for common parameters), and (iii) parameter estimation (for model-specific parameters). For \textit{3D reconstruction}, we follow PlankAssembly~\cite{HuZZYYZ23} to report the precision, recall, and F1 score. Specifically, we use Hungarian matching to match the 3D bounding box of predicted primitives with the ground truth. A prediction is regarded as a true positive if its 3D intersection-over-union (IOU) with the ground truth is $>0.5$. For \textit{model retrieval}, we compare each predicted model ID with that of the corresponding ground truth primitive, and report the percentage of matches. For \textit{parameter estimation}, we compute the accuracy on all correctly retrieved models. A successful estimation means that all parameters are correct.

\subsection{Experiments on the Input}

\noindent \textbf{Effect of the image encoder.} We first demonstrate that modern vision models (\ie, ViT) can understand the engineering drawing. For this experiment, we directly modify the PlankAssembly model~\cite{HuZZYYZ23}, which takes 2D lines from three orthographic views as input, and produces a 3D model consisting of 3D axis-aligned bounding boxes (\ie, planks). To enable PlankAssembly to take image as input, we remove the Transformer encoder (12M) that was used to encode vectorized input, and integrate a TinyViT~\cite{WuZPLXFY22}\footnote{\url{https://huggingface.co/timm/tiny_vit_21m_512.dist_in22k_ft_in1k}} with comparable model size (21M) to encode the raster image. The decoder then takes the visual tokens generated by TinyViT as input and outputs the 3D model.

To train the modified PlankAssembly model, we use the same code and dataset from its website\footnote{\url{https://github.com/manycore-research/PlankAssembly}}. For the input raster images, we generate an engineering drawing by arranging the three axis-aligned views on a fixed-size canvas. We also inject noises into the drawings during training and testing, as the original method does. 

\begin{figure}[t]
    \centering
    \includegraphics[width=0.8\linewidth]{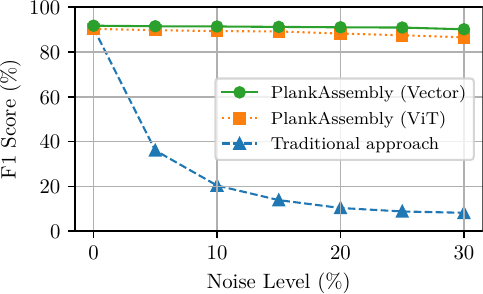}
    \caption{Effect of the image encoder. Results of PlankAssembly (Vector) and the traditional approach are directly taken from~\cite{HuZZYYZ23}.}
    \label{fig:input:modality}
\end{figure}

\cref{fig:input:modality} reports the F1 scores on varying input noise levels. As we can see, the model with the raster image input performs comparably to the model with vectorized input. This suggests that a modern general-purpose vision model is as effective as a domain-specific encoder.

\smallskip 
\noindent \textbf{Effect of the annotation layer.} One greatest advantage of using an image encoder is that it imposes fewer restrictions on the input as prior methods do, which typically assume that the drawing consists of exactly three orthographic views in vector-graphics format, and that the annotation layer is removed. In this experiment, we investigate the impact of the annotation layer on the reconstruction task. 

For this experiment, we use our new dataset because the PlankAssembly dataset does not provide any annotation. We compare the performance of \method model trained with two different inputs: (i) geometry layer only, and (ii) both geometry and annotation layers.

As shown in~\cref{tab:input:layer}, the model taking the original drawing (with both layers) as input achieves significantly higher accuracies across all metrics. Notably, this is in contrast to traditional practice which treats the annotation layer as nuisances that could hurt the performance of a reconstruction method. To gain insight into the performance gain, we show example programs generated by these two models in \cref{lst:layer}. As one can see, the dimension sets in the annotation provide direct reference \wrt the size of the primitives, which may be exploited by a vision-language model. Without the annotation, the model must infer the sizes based on the geometric entities, resulting in less accurate predictions.

\begin{table}[t]
    \centering
    \setlength{\tabcolsep}{2pt}
    \begin{tabular}{cc|ccccc}
        \toprule
        \multicolumn{2}{c|}{input layers} & retrieval & \multicolumn{3}{c}{reconstruction} & param. \tabularnewline
        geometry & annotation & acc. & prec. & rec. & F1 & acc. \tabularnewline
        \midrule
        \checkmark & & 85.42 & 63.00 & 62.48 & 62.65 & 81.94 \tabularnewline
        \checkmark & \checkmark & 93.80 & 83.10 & 82.56 & 82.76 & 97.21 \tabularnewline
        \bottomrule
    \end{tabular}
    \caption{Effect of the annotation layer.}
    \label{tab:input:layer}
\end{table}

\begin{table*}[t]
    \centering
    \begin{tabular}{c|ccc|ccccc}
        \toprule
        \multicolumn{4}{c|}{output} & retrieval & \multicolumn{3}{c}{reconstruction} & param. \tabularnewline
        format & model ID & common param. & model-specific param. & acc. & prec. & rec. & F1 & acc. \tabularnewline
        \midrule
        command & \checkmark & \checkmark & & 93.36 & 82.46 & 81.86 & 82.07 & n/a \tabularnewline
        \midrule
        text & \checkmark & \checkmark & & 93.84 & 82.94 & 82.51 & 82.65 & n/a \tabularnewline
        text & \checkmark & \checkmark & \checkmark & 93.80 & 83.10 & 82.56 & 82.76 & 97.21 \tabularnewline
        \bottomrule
    \end{tabular}
    \caption{Experimental results on output representation.}
    \label{tab:output}
\end{table*}

\subsection{Experiments on the Output}

Next, we study the performance of our model with different ouput representations. To verify the effectiveness of the proposed text-based shape program, we compare it with a domain-specific sequence representation. As discussed in~\cref{sec:program}, a method using domain-specific sequence representation with a command template lacks the flexibility to predict model-specific parameters. Thus, for this experiment, we compare \method to two variants which predict the model ID and common parameters only: 

\begin{itemize}
    \item The \emph{first variant} outputs domain-specific sequence representations. To this end, we create a command template that includes the model ID and all common parameters. For common parameters, we quantize each position and size parameter into 1500 bins with a resolution of 3mm, and the rotation angle into 4 bins. For model ID, we use the same special token as described in~\cref{sec:model}. This variant has a standard encoder-decoder architecture, including InternViT-300M as the encoder and a 6-layer Transformer decoder (similar to the decoder used in PlankAssembly~\cite{HuZZYYZ23}).

    \item The \emph{second variant} has the same architecture as \method, but is trained to predict a Python-based shape program describing the model ID and common parameters only.
\end{itemize}

\begin{figure*}[t]
    \centering
    \includegraphics[width=0.95\linewidth]{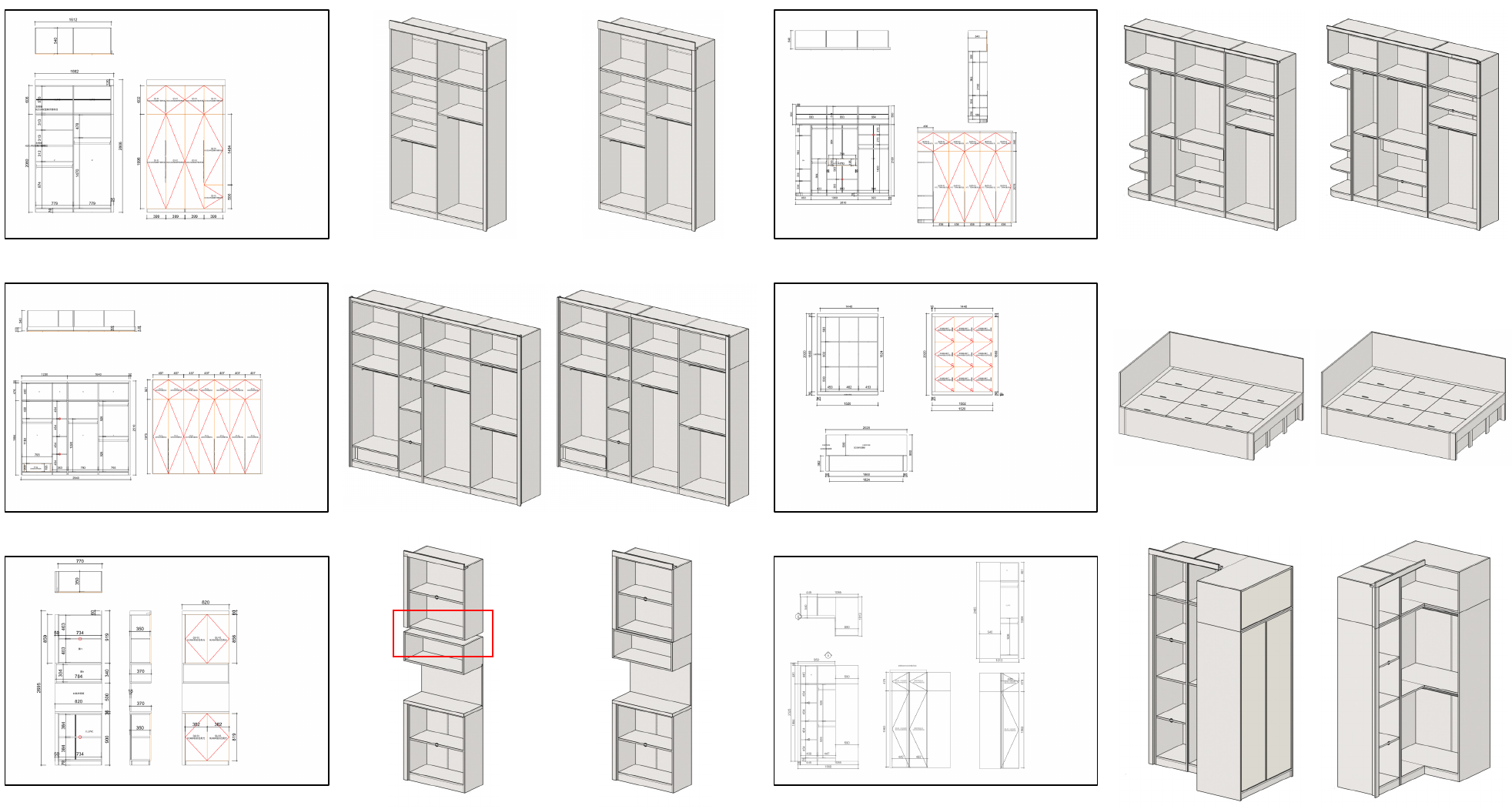}
    \caption{Qualitative results. For each case, we show {\bf from left to right} the input drawing, the model reconstructed by \method, and the ground truth. To reveal the inside structure, we manually delete the doors in the visualization.}
    \label{fig:visuals}
\end{figure*}

The results are shown in~\cref{tab:output}. We observe that all three models show comparable performance on both model retrieval and 3D reconstruction. This is significant because it suggests that using free-form text as output is as effective as employing a domain-specific sequence representation as in prior work (\eg, PlankAssembly). Moreover, our \method model with text-based shape program offers greater flexibility in designing the output sequence and performs well across all metrics, \ie, 3D reconstruction, model retrieval, and model-specific parameter estimation.

\cref{fig:visuals} visualizes some 3D parametric reconstruction results of \method. As one can see, our method successfully reconstructs a wide variety of cabinet models. Note that the input 2D drawings here would cause major problems for previous methods (\eg, PlankAssembly) for several reasons. \emph{First}, they contain a varying number of orthographic views. For example, in Figure~\ref{fig:visuals} (first row, left), the drawing does not show the side view  (instead, a section view is provided to better reveal the inside design), whereas in~\cref{fig:visuals} (third row, right), five views are shown. \emph{Second}, the different views in a drawing may not be aligned to each other. \emph{Third}, both the geometry layer and annotation layer are present. 

In the last row of~\cref{fig:visuals}, we illustrate some common artifacts in the models reconstructed by our method. On the left, we show a case where our method produces a model consisting of two parts, with a small gap in between (highlighted by the red box). This suggests that the predicted 3D positions (\ie, $p_i$) may not be accurate enough. Note that, unlike the sizes (\ie, $s_i$) which can often be directly referred from the annotations, 3D positions need to be inferred, sometimes through arithmetic calculations. This remains a challenging task for LLM. On the right, due to ambiguity in the coordinate system used, the L-shape cabinet is reconstructed with a different pose \wrt the ground truth.

\begin{table}[t]
    \centering
    \setlength{\tabcolsep}{3pt}
    \begin{tabular}{c|ccccc}
        \toprule
        \multirow{2}{*}{proxy language} & retrieval & \multicolumn{3}{c}{reconstruction} & param. \tabularnewline
        & acc. & prec. & rec. & F1 & acc. \tabularnewline
        \midrule
        YAML & 93.12 & 83.18 & 83.07 & 83.05 & 97.09 \tabularnewline
        Python & 93.80 & 83.10 & 82.56 & 82.76 & 97.21 \tabularnewline
        \bottomrule
    \end{tabular}
    \caption{Ablation on proxy language.}
    \label{tab:language}
\end{table}

\smallskip
\noindent \textbf{Ablation on the proxy language.} To examine the impact of our choice of proxy language, we have conducted an experiment in which we replace Python with YAML, a popular data serialization language. Compared to Python, the YAML script is generally longer. Specifically, the average script lengths for 3D models in our dataset are 697 and 1208 tokens for Python and YAML, respectively.

As shown in \cref{tab:language}, the models trained with both languages yield very similar results, indicating that the choice of proxy language does not significantly affect the model performance for our task.

\section{Conclusion}

In this paper, we present a vision-language approach to 3D parametric model reconstruction from 2D CAD drawings. Our key observations are: (i) a modern vision encoder (\eg, ViT) is effective in understanding CAD drawings, and (ii) using text-based sequence representation provides greater flexibility in incorporating a large number of primitives with arbitrary geometric, semantic, and functional properties. 

The studies presented in this paper are limited to cabinet furniture only. However, we point out that our method is general and can be applied to other types of CAD models -- if a large-scale dataset for the domain of interest is available. Meanwhile, the use of vision-language foundation models opens up many directions for future research. A direction of particular interest is visual question answering in the context of 2D CAD drawings, a critical capacity for developing autonomous agents for design and manufacturing.

\appendix

\section*{Acknowledgements} This work was supported by the National Science and Technology Major Project (No. 2022ZD0117800), Young Elite Scientists Sponsorship Program by CAST, and the Fundamental Research Funds for the Central Universities.

\bibliography{aaai25}

\newpage

In the appendix, we provide additional training details in \cref{sec:detail}, inference details of \method in \cref{sec:inference}, and additional ablation study in \cref{sec:ablation}.

\section{Additional Training Details}
\label{sec:detail}

\subsection{Experiments on the Input} 

\begin{figure}[h]
    \centering
    \fboxsep=0pt
    \setlength{\tabcolsep}{1pt}
    \begin{tabular}{cc}
        \fbox{\includegraphics[width=0.45\linewidth]{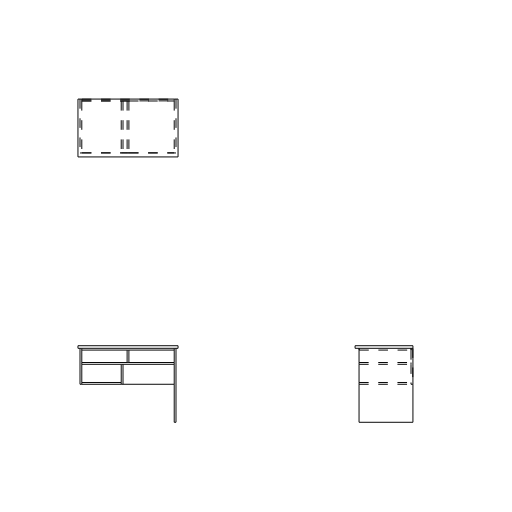}} &
        \fbox{\includegraphics[width=0.45\linewidth]{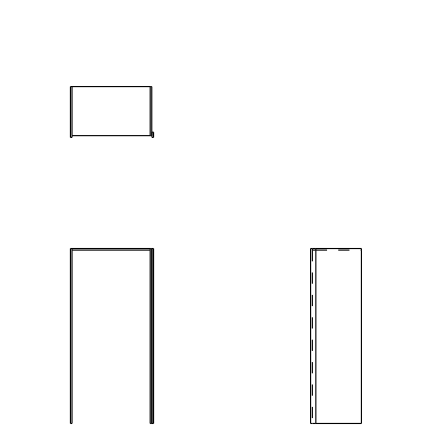}} \tabularnewline
    \end{tabular}
    \caption{Examples of the generated engineering drawings. For each case, the top, front, and side views are placed at the top left, bottom left, and bottom right of the canvas, respectively.}
    \label{fig:plankassembly}
\end{figure}

\noindent \textbf{Effect of the image encoder.} In this experiment, we develop a variant of PlankAssembly~\cite{HuZZYYZ23}, called PlankAssembly (ViT), which takes a raster image as input. To train the model, we generate engineering drawings by arranging the three orthographic views provided in the original PlankAssembly dataset on a fixed-size canvas. The resolution of the canvas is set to $512 \times 512$. \cref{fig:plankassembly} shows some examples.

For training \textit{PlankAssembly (ViT)}, we use the same training configurations as \textit{PlankAssembly (Vector)}. We train it for 400K iterations using Adam optimizer~\cite{KingmaB15} with a learning rate of $10^{-4}$ and a total batch size of 64.

\smallskip
\noindent \textbf{Effect on the annotation layer.}  
In this paper, we train \method with two types of input: (i) geometry layer only, and (ii) both geometry and annotation layers. \cref{fig:layers} shows an example for the two types of input.

\begin{figure}[h]
    \centering
    \fboxsep=0pt
    \begin{tabular}{cc}
        \fbox{\includegraphics[width=0.45\linewidth]{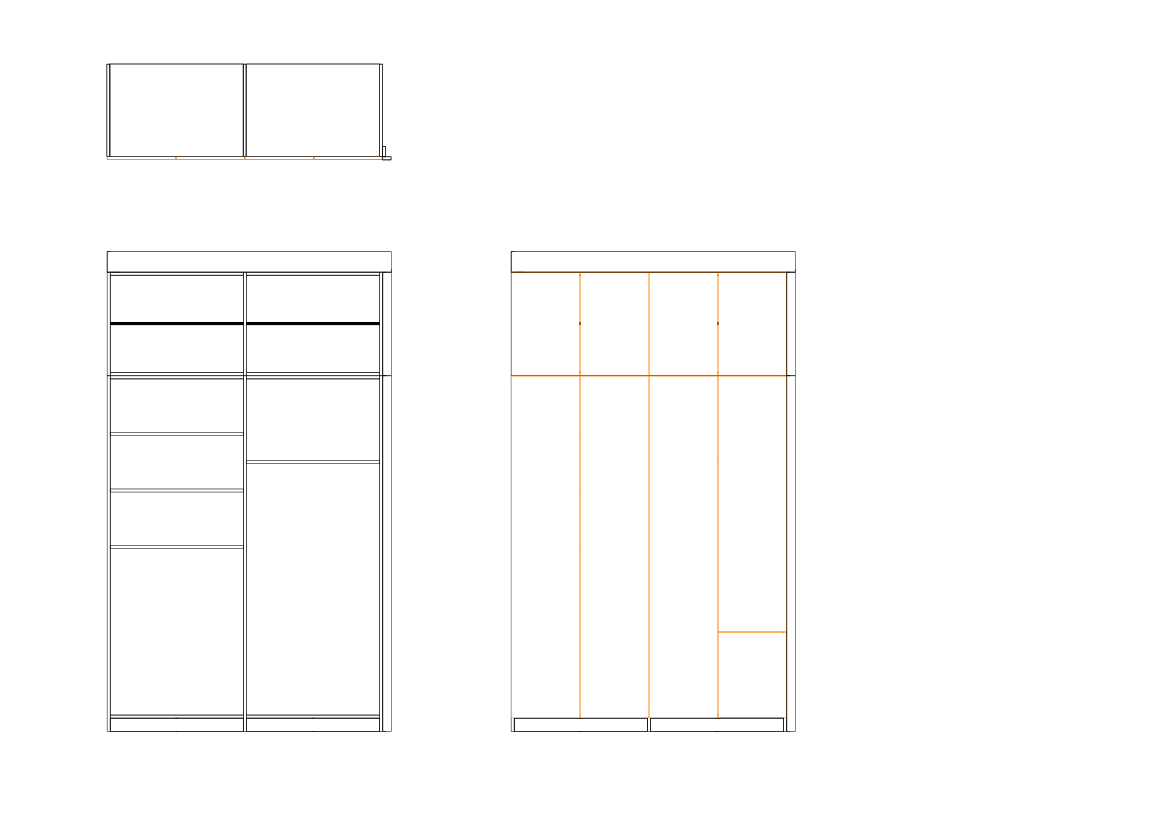}} &
        \fbox{\includegraphics[width=0.45\linewidth]{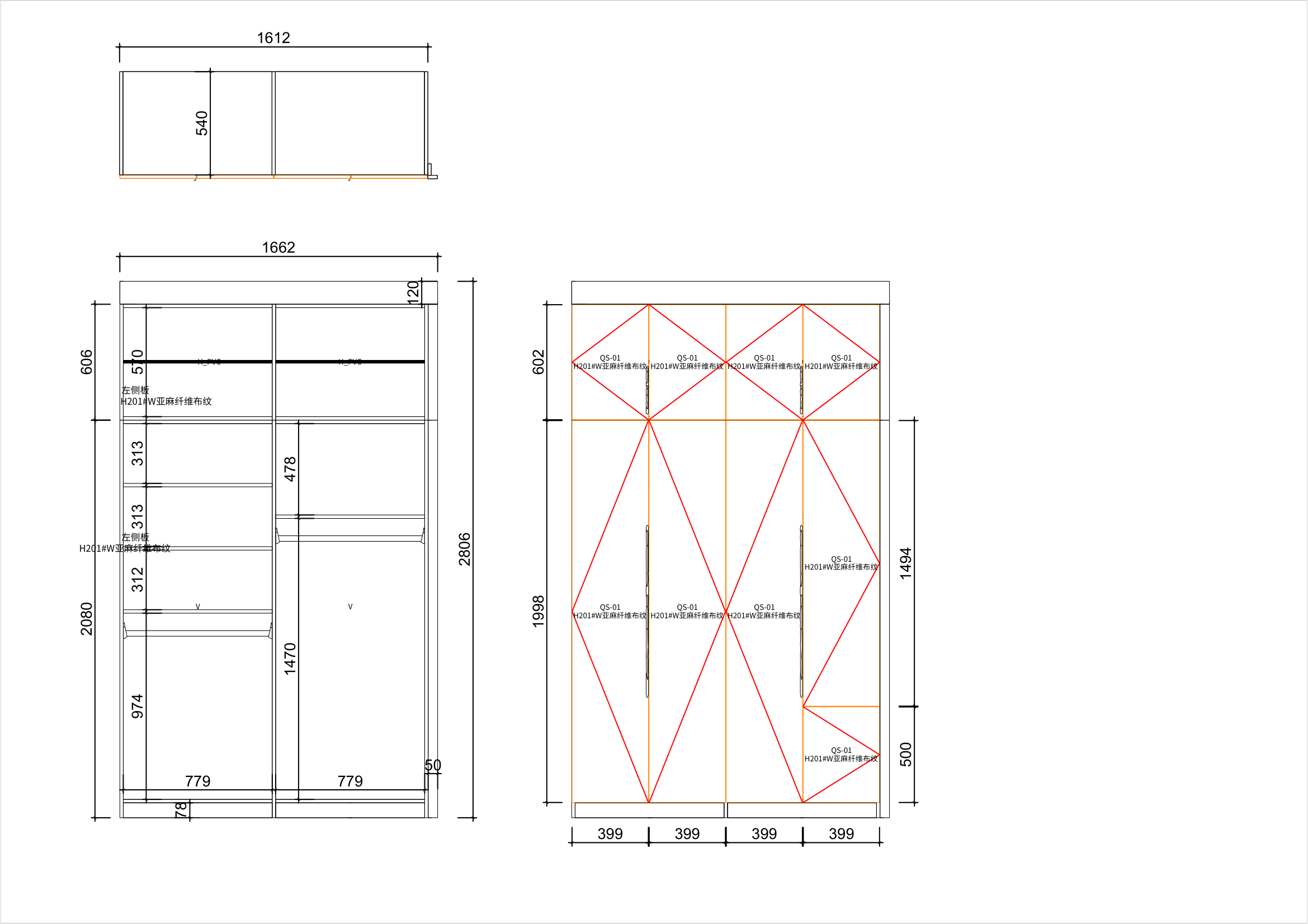}}
        \tabularnewline
    \end{tabular}
    \caption{An example CAD drawing. {\bf Left:} geometry layer only. {\bf Right:} both the geometry and annotation layers}.
    \label{fig:layers}
\end{figure}

\subsection{Experiments on the Output}

In this experiment, we train two variants of \method.

For the \textit{first variant} which outputs domain-specific sequence representations, we train it for 56K iterations using 32 NVIDIA RTX 4090 GPU devices. We use the AdamW optimizer~\cite{LoshchilovH17} with a learning rate of $5 \times 10^{-5}$. The total batch size is 128.

The \textit{second variant} is trained using the same configuration as \method.

\section{Inference Details of \method}
\label{sec:inference}

At the inference time, we use the greedy sampling strategy and limit \method to generate a maximum number of 4096 tokens. 

\begin{figure*}[t]
    \begin{tcolorbox}[
        width=\linewidth, 
        colback=white, colframe=black, 
        left=10pt, right=10pt, top=10pt, bottom=10pt]

        \begin{minipage}[ft]{0.5\linewidth} 
            {\color{prompt}\textbf{Prompt:}} {\color{img}\textbf{<img>}} Reconstruct cabinet from image. The bounding box reference code is as follows:
\begin{lstlisting}
@dataclasses.dataclass
class Bbox:
    position_x: float
    position_y: float
    position_z: float
    scale_x: float
    scale_y: float
    scale_z: float
    angle_z: float
\end{lstlisting}
        \end{minipage}
        \hfill
        \begin{minipage}[f]{0.5\linewidth}
            \centering
            \fboxsep=0pt
            \fbox{\includegraphics[width=0.8\linewidth]{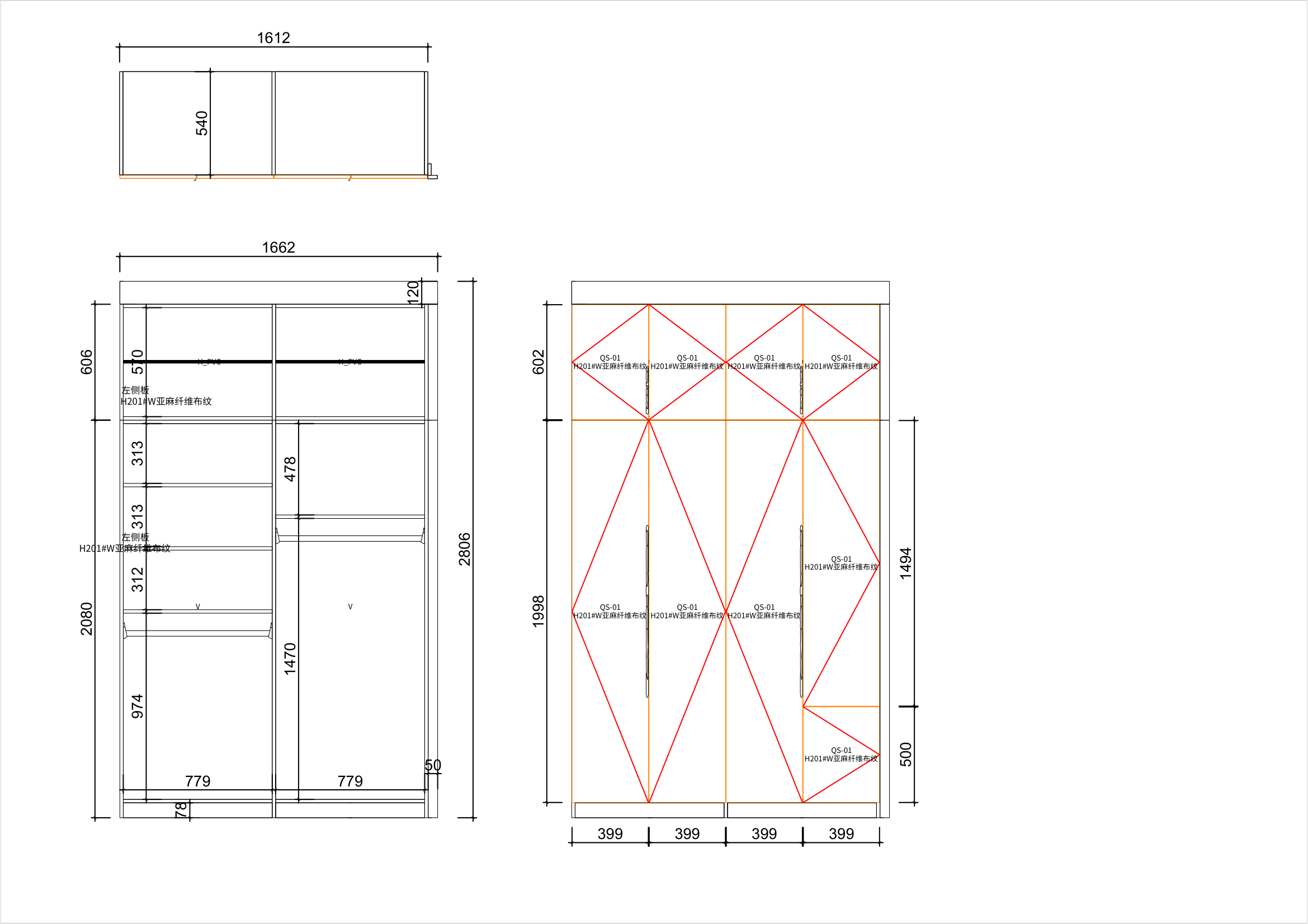}}
        \end{minipage}
        \\\\\\
        \begin{minipage}[ft]{\linewidth}
            {\color{prompt}\textbf{Response:}}
\begin{lstlisting}
bbox_0 = Bbox(1637, 280, 1040, 50, 560, 2080, 0)
model_0 = <model_57763823>()
bbox_1 = Bbox(1637, 280, 2383, 50, 560, 606, 0)
model_1 = <model_57763823>()
bbox_2 = Bbox(806, 290, 1040, 1612, 540, 2080, 0)
model_2 = <model_132924318>(N=2, NKA=779, NKB=779, BT=9)
bbox_3 = Bbox(407.5, 280, 1079, 779, 520, 18, 0)
model_3 = <model_76969651>()
bbox_4 = Bbox(407.5, 280, 978.5, 779, 40, 83, 0)
model_4 = <model_57400210>()
bbox_5 = Bbox(407.5, 280, 1740, 779, 520, 18, 0)
model_5 = <model_76969651>()
bbox_6 = Bbox(407.5, 280, 1409, 779, 520, 18, 0)
model_6 = <model_76969651>()
bbox_7 = Bbox(1204.5, 280, 1575, 779, 520, 18, 0)
model_7 = <model_76969651>()
bbox_8 = Bbox(1204.5, 280, 1474.5, 779, 40, 83, 0)
model_8 = <model_57400210>()
bbox_9 = Bbox(806, 290, 2383, 1612, 540, 606, 0)
model_9 = <model_115813862>(N=2, NKA=779, NKB=779, BT=9)
bbox_10 = Bbox(831, 280, 2746, 1662, 560, 120, 0)
model_10 = <model_57761062>()
\end{lstlisting}
        \end{minipage}
    \end{tcolorbox}
    \caption{A example conversation using Python.}
    \label{fig:python}
\end{figure*}

\begin{figure*}[t]
    \begin{tcolorbox}[
        width=\linewidth,
        colback=white, colframe=black,
        left=10pt, right=10pt, top=10pt, bottom=10pt]

        \begin{minipage}[ft]{0.5\linewidth} 
            {\color{prompt}\textbf{Prompt:}} {\color{img}\textbf{<img>}} Reconstruct cabinet from image.
        \end{minipage}
        \hfill
        \begin{minipage}[f]{0.5\linewidth}
            \centering
            \fboxsep=0pt
            \fbox{\includegraphics[width=0.8\linewidth]{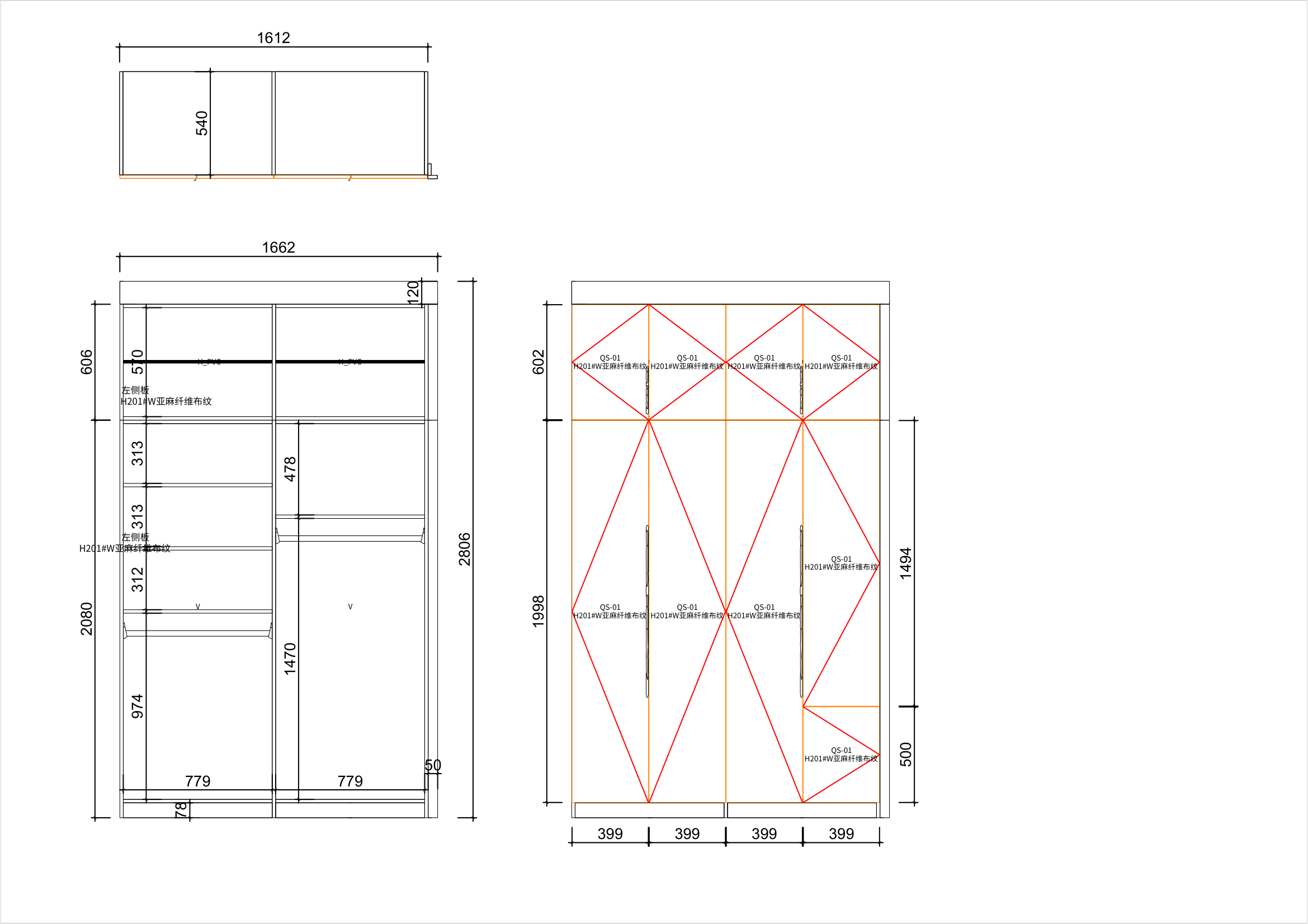}}
        \end{minipage}
        \\\\\\
        \begin{minipage}[ft]{\linewidth}
            {\color{prompt}\textbf{Response:}}
\begin{lstlisting}
- id: 0
  position_x: 1637
  position_y: 280
  position_z: 1040
  scale_x: 50
  scale_y: 560
  scale_z: 2080
  angle_z: 0
  model_id: <model_57763823>
- id: 1
  position_x: 1637
  position_y: 280
  position_z: 2383
  scale_x: 50
  scale_y: 560
  scale_z: 606
  angle_z: 0
  model_id: <model_57763823>
$\vdots$
- id: 9
  position_x: 806
  position_y: 290
  position_z: 2383
  scale_x: 1612
  scale_y: 540
  scale_z: 606
  angle_z: 0
  model_id: <model_115813862>
  N: 2
  NKA: 779
  NKB: 779
  BT: 9
- id: 10
  position_x: 831
  position_y: 280
  position_z: 2746
  scale_x: 1662
  scale_y: 560
  scale_z: 120
  angle_z: 0
  model_id: <model_57761062>
\end{lstlisting}
        \end{minipage}
    \end{tcolorbox}
    \caption{A example conversation using YAML.}
    \label{fig:yaml}
\end{figure*}

\cref{fig:python,fig:yaml} shows the full prompt and an example response in Python and YAML format, respectively.

\section{Ablation on Foundation Model}
\label{sec:ablation}

To demonstrate that our approach can adapt to other foundation models, we train with the QWen2-VL-2B model~\cite{Qwen2VL}\footnote{\url{https://huggingface.co/Qwen/Qwen2-VL-2B-Instruct}} in this experiment. We follow the same training setting as Mini-InternVL-1.5-2B. Note that this VLM has been trained extensively on a diverse dataset (including OCR data). As shown in \cref{tab:qwenvl}, QWen2-VL-2B achieves similar performance on model retrieval and parameter estimation, and is slightly better on 3D box reconstruction. This is expected as QWen2-VL-2B demonstrates superior performance on various VLM benchmarks.

\begin{table}[h]
    \centering
    \setlength{\tabcolsep}{2pt}
    \begin{tabular}{c|ccccc}
        \toprule
        \multirow{2}{*}{VLM} & retrieval & \multicolumn{3}{c}{reconstruction} & param. \tabularnewline
        & acc. & prec. & rec. & F1 & acc. \tabularnewline
        \midrule
        Mini-InternVL-1.5-2B & 93.80 & 83.10 & 82.56 & 82.76 & 97.21 \tabularnewline
        QWen2-VL-2B & 93.98 & 85.34 & 84.60 & 84.90 & 96.93\tabularnewline
        \bottomrule
    \end{tabular}
    \caption{Ablation study on the VLMs.}
    \label{tab:qwenvl}
\end{table}

\end{document}